\documentclass[10pt]{article}
\usepackage[utf8]{inputenc}
\usepackage[T1]{fontenc}
\usepackage{amsmath}
\usepackage{amsthm}
\usepackage{amsfonts}
\usepackage{amssymb}
\usepackage{makeidx}
\usepackage{graphicx}
\usepackage{algorithm2e}
\usepackage{lmodern}
\usepackage{mathtools}
\usepackage{url}
\usepackage{xcolor}
\usepackage[left=2cm,right=2cm,top=1cm,bottom=2cm]{geometry}
\usepackage{tikz}
\usetikzlibrary{positioning, arrows.meta, shapes.geometric}
\usepackage{multirow}
\usepackage{caption}
\usepackage{subcaption}
\usepackage{tabularx} 
\usepackage{booktabs}

\author{Anna Niane \thanks{African Institute for Mathematical Sciences (AIMS), Mbour, Senegal.
{anna.niane@aims-senegal.org}}
\and Prudence Djagba \thanks{Lyman Briggs College, Michigan State University, East Lansing, Michigan, USA.
{djagbapr@msu.edu}}}
\title{Graph-Theoretic Models for the Prediction of Molecular Measurements}

\usepackage{hyperref}
\usepackage{cleveref}

\theoremstyle{remark}

\theoremstyle{definition}

\date{}

\begin{document}
\maketitle	
\begin{abstract}
\noindent
Graph-theoretic approaches offer simplicity, interpretability, and low computational cost for molecular property prediction. Among these, the model proposed by Mukwembi and Nyabadza, based on the external activity $D(G)$ and internal activity $\zeta(G)$ indices, achieved strong results on a small flavonoid dataset. However, its ability to generalize to larger and chemically diverse datasets has not been tested. This study evaluates the baseline $D(G)$-$\zeta(G)$ polynomial model on five benchmark datasets from MoleculeNet, covering biological activity (BACE, 1,513 molecules), lipophilicity (LogP synthetic, 14,610 molecules; LogP experimental, 753 molecules), aqueous solubility (ESOL, 1,128 molecules), and hydration free energy (SAMPL, 642 molecules). The baseline model achieves an average $R^2 = 0.24$, confirming limited transferability. To address this, a systematic enhancement framework is proposed, progressively incorporating Ridge regularization, additional graph descriptors, physicochemical properties, ensemble learning with Gradient Boosting, Lasso feature selection, and a hybrid approach combining topological indices with Morgan fingerprints. The enhanced models raise the average best $R^2$ to 0.79, with individual improvements ranging from 165\% to 274\%. All improvements are statistically significant ($p < 0.001$). A direct comparison with a Graph Convolutional Network under identical experimental conditions shows that the enhanced classical models match or outperform deep learning on all five datasets. Comparison with the recent GNN+PGM hybrid of Djagba et al.\ further confirms competitiveness, with the enhanced models achieving the best results on two datasets and tying on one. The entire framework requires no GPU, trains in under five minutes, and uses only open-source tools, making it accessible for researchers in resource-limited settings.

\end{abstract}
\textbf{Keywords:} Molecular property prediction, topological indices, graph-theoretic models, QSAR, machine learning, enhancement framework, molecular fingerprints, drug discovery, MoleculeNet.

\section{Introduction}

Developing new drugs is one of the most expensive challenges in modern science. It takes over a decade, costs billions of dollars, and most candidate molecules fail along the way. To reduce this cost, computational methods have become essential for predicting molecular properties before compounds are synthesized and tested in the laboratory.

These predictions rely on a simple principle: the structure of a molecule determines its properties. This idea is at the heart of Quantitative Structure-Property Relationship (QSPR) and Quantitative Structure-Activity Relationship (QSAR) modeling, which use mathematical relationships between structural features and observed properties such as solubility, lipophilicity, and binding affinity \cite{hansch1964correlation}.

Graph theory provides a natural way to represent molecules. Atoms become vertices and bonds become edges, forming a molecular graph. From this graph, numerical descriptors called topological indices can be computed. These indices capture molecular size, shape, and connectivity without requiring three-dimensional coordinates. The use of topological indices for property prediction dates back to Wiener's 1947 work on alkane boiling points \cite{wiener1947structural}. Since then, many indices have been developed, including the Zagreb indices \cite{gutman1972graph} and the Randi\'{c} connectivity index \cite{randic1975characterization}.

Recently, Mukwembi and Nyabadza \cite{mukwembi2023flavonoids} introduced a model that uses two novel graph invariants: the external activity $D(G)$ and the internal activity $\zeta(G)$. Their polynomial regression model achieved $R^2 = 1$ on a dataset of 9 flavonoid compounds, which is a remarkable result. However, modern drug discovery operates at much larger scales with chemically diverse compounds, and the transferability of this model to such settings has not been investigated.

At the same time, deep learning methods such as Graph Neural Networks (GNNs) have gained popularity for molecular property prediction \cite{gilmer2017neural, kipf2017semi}. GNNs learn molecular representations directly from graph structures and have shown competitive performance on standard benchmarks \cite{wu2018moleculenet}. However, they require significant computational resources, specialized expertise, and function as black-box models with limited interpretability.

This raises a practical question: can classical graph-theoretic models, when systematically enhanced, achieve competitive performance without these costs? This study addresses this question through three contributions. First, we provide the first large-scale evaluation of the Mukwembi-Nyabadza model on five diverse benchmark datasets. Second, we develop a systematic enhancement framework that progressively improves the baseline model. Third, we directly compare the enhanced classical models with a Graph Convolutional Network (GCN) and with the published deep learning results of Djagba et al.\ \cite{djagba2025hybrid}, who used the same benchmark datasets.

\section{Background of the Study}

\subsection{Molecular quantities in drug discovery}

Four molecular quantities are central to this study. The half-maximal inhibitory concentration (IC$_{50}$) measures how strongly a compound inhibits a biological target. In practice, it is often expressed as pIC$_{50} = -\log_{10}(\text{IC}_{50})$, where lower IC$_{50}$ values indicate more potent drugs. The partition coefficient (LogP) measures how much a molecule prefers fatty environments over water:
\begin{equation*}
\text{LogP} = \log_{10} \frac{[\text{Compound}]_{\text{octanol}}}{[\text{Compound}]_{\text{water}}}
\end{equation*}
Aqueous solubility (LogS) describes how much of a compound dissolves in water, which directly affects oral bioavailability. Hydration free energy ($\Delta G_{\text{hyd}}$) measures how favorable it is to transfer a molecule from the gas phase into water, providing a thermodynamic measure relevant to binding affinity and solubility.

\subsection{The Mukwembi-Nyabadza model}

The baseline model uses two graph invariants computed from the molecular graph $G = (V, E)$, where $V$ represents atoms and $E$ represents bonds. The vertices are classified as external (degree $= 1$) or internal (degree $\geq 2$).

For each vertex $v$, a score function $s(v)$ is defined. For external vertices, $s(v) = \sum_{u \in V_{\text{ext}}} d(u,v)$, where $d(u,v)$ is the shortest path distance. For internal vertices, $s(v) = \frac{t(v)}{\deg(v)} \cdot \varepsilon(v)$, where $\varepsilon(v)$ is the eccentricity and $t(v)$ is the irregularity index.

The external activity and internal activity are then:
\begin{equation*}
D(G) = \frac{\sum_{v \in V_{\text{ext}}} s(v)}{n^3}, \qquad \zeta(G) = \frac{\sum_{v \in V_{\text{int}}} s(v)}{n^2}
\end{equation*}
where $n = |V|$ is the number of vertices. The polynomial regression model combines both indices:
\begin{equation*}
\hat{y}(G) = \alpha_1 + \alpha_2 D + \alpha_3 \zeta + \alpha_4 D^2 + \alpha_5 D\zeta + \alpha_6 \zeta^2 + \alpha_7 D^2\zeta + \alpha_8 D\zeta^2 + \alpha_9 \zeta^3
\end{equation*}
where $\alpha_1, \ldots, \alpha_9$ are fitted using ordinary least squares.

\subsection{Deep learning approaches}

Graph Neural Networks represent molecules as graphs where atoms are nodes and bonds are edges. Through message-passing operations, each atom builds a representation by aggregating information from its neighbors \cite{gilmer2017neural}. The Graph Convolutional Network (GCN) of Kipf and Welling \cite{kipf2017semi} is one of the standard architectures for this task.

Djagba et al.\ \cite{djagba2025hybrid} recently proposed a hybrid framework combining GNNs with Probabilistic Graphical Models (GNN+PGM) for molecular property prediction. Their study is directly relevant to this work because it uses the same benchmark datasets from MoleculeNet and explicitly builds upon the Mukwembi-Nyabadza model. Their hybrid approach achieved $R^2$ values ranging from 0.66 to 0.91, but requires an NVIDIA A100 GPU and specialized deep learning expertise.

\section{Implementation, Results, and Discussions}

\subsection{Implementation}

\subsubsection{Datasets}

Five benchmark datasets from MoleculeNet \cite{wu2018moleculenet} were used to evaluate the enhancement framework. Table~\ref{tab:datasets} summarizes their characteristics.

\begin{table}[ht!]
\centering
\begin{tabular}{|l|c|c|c|c|}
\hline
\textbf{Dataset} & \textbf{Size} & \textbf{Target} & \textbf{Units} & \textbf{Property Type} \\
\hline
BACE & 1,513 & pIC$_{50}$ & -- & Biological activity \\
\hline
LogP Synthetic & 14,610 & LogP & -- & Physicochemical \\
\hline
LogP Experimental & 753 & LogP & -- & Physicochemical \\
\hline
ESOL & 1,128 & LogS & log(mol/L) & Physicochemical \\
\hline
SAMPL & 642 & $\Delta G_{\text{hyd}}$ & kcal/mol & Thermodynamic \\
\hline
\end{tabular}
\caption{Benchmark datasets used in this study.}
\label{tab:datasets}
\end{table}

These datasets cover four different molecular properties and range from 642 to 14,610 molecules, providing a rigorous test of model generalization across varying dataset sizes and chemical diversity.

\subsubsection{Enhancement framework}

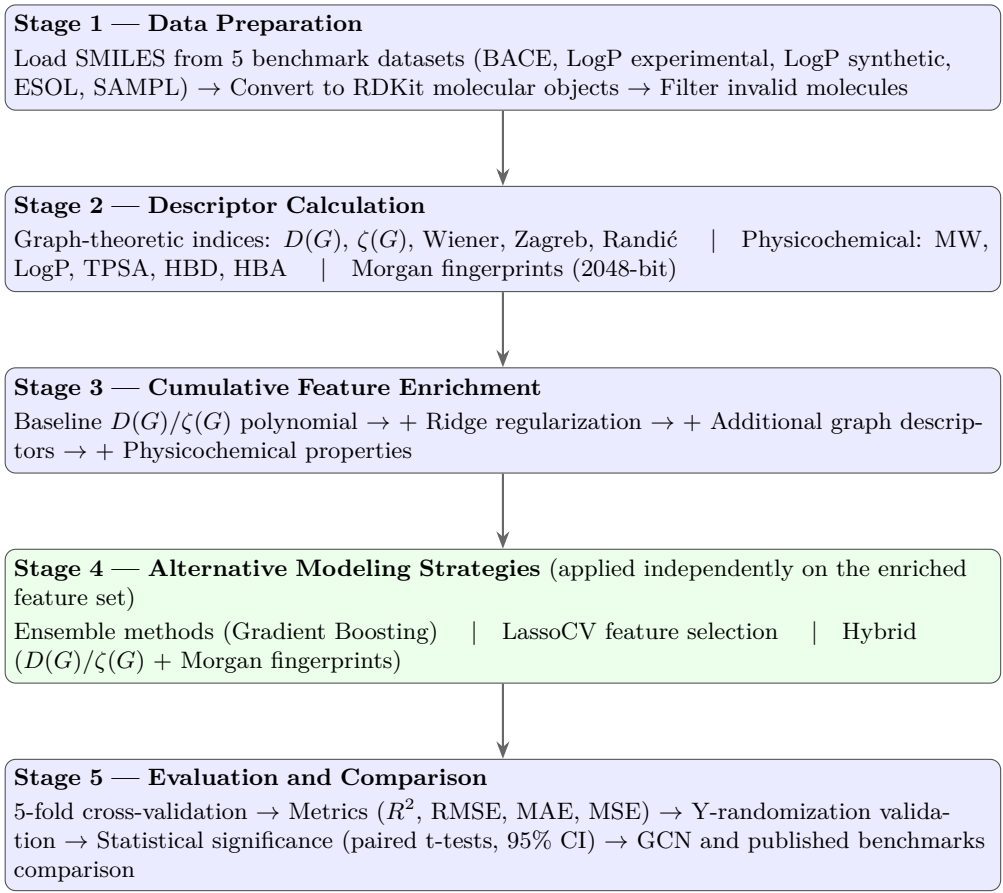
\begin{figure}[htbp]
\centering
\begin{tikzpicture}[
    node distance=1.0cm,
    stage/.style={
        rectangle, rounded corners, draw=black!70, fill=blue!8,
        text width=13cm, minimum height=1.4cm, align=left,
        font=\small
    },
    branch/.style={
        rectangle, rounded corners, draw=black!70, fill=green!8,
        text width=13cm, minimum height=1.4cm, align=left,
        font=\small
    },
    arrow/.style={-Stealth, thick, draw=black!60},
    note/.style={font=\footnotesize\itshape, text=black!60}
]

\node[stage] (S1) {
    \textbf{Stage 1 --- Data Preparation}\\[2pt]
    Load SMILES from 5 benchmark datasets (BACE, LogP experimental,
    LogP synthetic, ESOL, SAMPL) $\rightarrow$ Convert to RDKit
    molecular objects $\rightarrow$ Filter invalid molecules
};

\node[stage, below=of S1] (S2) {
    \textbf{Stage 2 --- Descriptor Calculation}\\[2pt]
    Graph-theoretic indices: $D(G)$, $\zeta(G)$, Wiener, Zagreb,
    Randi\'{c} \quad$|$\quad
    Physicochemical: MW, LogP, TPSA, HBD, HBA \quad$|$\quad
    Morgan fingerprints (2048-bit)
};

\node[stage, below=of S2] (S3) {
    \textbf{Stage 3 --- Cumulative Feature Enrichment}\\[2pt]
    Baseline $D(G)$/$\zeta(G)$ polynomial
    $\rightarrow$ + Ridge regularization
    $\rightarrow$ + Additional graph descriptors
    $\rightarrow$ + Physicochemical properties
};

\node[branch, below=of S3] (S4) {
    \textbf{Stage 4 --- Alternative Modeling Strategies}
    (applied independently on the enriched feature set)\\[2pt]
    Ensemble methods (Gradient Boosting) \quad$|$\quad
    LassoCV feature selection \quad$|$\quad
    Hybrid ($D(G)$/$\zeta(G)$ + Morgan fingerprints)
};

\node[stage, below=of S4] (S5) {
    \textbf{Stage 5 --- Evaluation and Comparison}\\[2pt]
    5-fold cross-validation $\rightarrow$
    Metrics ($R^2$, RMSE, MAE, MSE) $\rightarrow$
    Y-randomization validation $\rightarrow$
    Statistical significance (paired t-tests, 95\% CI) $\rightarrow$
    GCN and published benchmarks comparison
};

\draw[arrow] (S1) -- (S2);
\draw[arrow] (S2) -- (S3);
\draw[arrow] (S3) -- (S4);
\draw[arrow] (S4) -- (S5);

\node[note, right] at (S3.east) {};
\node[note, right] at (S4.east) {};

\end{tikzpicture}
\caption{Overview of the complete methodological pipeline.} 
\label{fig:pipeline}
\end{figure}

The core methodology of this study is a systematic enhancement framework that progressively adds new features and modeling techniques to the baseline $D(G)$-$\zeta(G)$ model. Figure~\ref{fig:pipeline} presents an overview of the complete methodological pipeline.

The framework consists of seven approaches organized in two phases. The first phase follows a cumulative progression. Approach 1 is the baseline polynomial model using only $D(G)$ and $\zeta(G)$. Approach 2 adds Ridge regularization ($L_2$ penalty) to improve coefficient stability:
\begin{equation*}
\hat{\beta}_{\text{Ridge}} = \arg\min_{\beta} \left\{ \sum_{i=1}^{n} (y_i - x_i^T \beta)^2 + \lambda \sum_{j=1}^{p} \beta_j^2 \right\}
\end{equation*}
Approach 3 adds classical graph-theoretic descriptors (Wiener index, Zagreb indices, Randi\'{c} index, Balaban index, graph diameter, radius, density, and clustering coefficient). Approach 4 adds physicochemical properties (molecular weight, topological polar surface area, hydrogen bond donors and acceptors, number of rotatable bonds, and aromaticity measures) computed using RDKit \cite{landrum2013rdkit} and Mordred \cite{moriwaki2018mordred}.

The second phase applies three independent modeling strategies on the enriched feature set from Approach 4. Approach 5 uses Gradient Boosting, an ensemble method that builds decision trees sequentially, where each tree corrects the errors of the previous one:
\begin{equation*}
F_m(x) = F_{m-1}(x) + \nu \cdot h_m(x)
\end{equation*}
where $\nu$ is the learning rate and $h_m$ is the $m$-th weak learner. Approach 6 uses Lasso regression for feature selection through $L_1$ regularization:
\begin{equation*}
\hat{\beta}_{\text{Lasso}} = \arg\min_{\beta} \left\{ \sum_{i=1}^{n} (y_i - x_i^T \beta)^2 + \lambda \sum_{j=1}^{p} |\beta_j| \right\}
\end{equation*}
Approach 7 is a hybrid that combines $D(G)$ and $\zeta(G)$ with 1,024-bit Morgan fingerprints \cite{rogers2010extended} using a Random Forest regressor.

\subsubsection{GCN implementation}

A GCN was implemented following Kipf and Welling \cite{kipf2017semi} to provide a direct comparison with deep learning. The architecture consists of three graph convolutional layers with 128 hidden units, batch normalization, ReLU activation, and dropout ($p = 0.2$). A global mean pooling operation converts atom-level representations into a molecule-level vector, followed by two fully connected layers. Training used the Adam optimizer with a learning rate of 0.001 and early stopping with a patience of 30 epochs. The GCN was evaluated under the exact same conditions as the classical models: same datasets, same 5-fold cross-validation splits, and same evaluation metrics.

\subsubsection{Evaluation protocol}

All models were evaluated using 5-fold cross-validation with four metrics: the coefficient of determination ($R^2$), root mean square error (RMSE), mean absolute error (MAE), and mean squared error (MSE). Statistical significance was assessed through paired $t$-tests comparing each enhanced model against the baseline, with 95\% confidence intervals. Y-randomization tests were performed to confirm that models capture genuine structure-property relationships rather than fitting noise.

\subsection{Results and Analysis}

\subsubsection{Baseline performance}

The baseline $D(G)$-$\zeta(G)$ model achieved $R^2$ values between 0.19 and 0.32 across the five datasets, with an average of 0.24. This represents a large drop from the $R^2 = 1$ reported on the original flavonoid dataset \cite{mukwembi2023flavonoids} and confirms that two topological indices alone are not sufficient for large, chemically diverse datasets.

\subsubsection{Enhancement results}

Table~\ref{tab:results_all} presents the $R^2$ values for all seven approaches on each dataset.

\begin{table}[ht!]
\centering
\small
\setlength{\tabcolsep}{4pt}
\begin{tabular}{|l|c|c|c|c|c|}
\hline
\textbf{Model} & \textbf{BACE} & \textbf{LogP syn.} & \textbf{LogP exp.} & \textbf{ESOL} & \textbf{SAMPL} \\
\hline
Baseline ($D/\zeta$) & 0.19 & 0.25 & 0.20 & 0.32 & 0.26 \\
\hline
+ Ridge & 0.19 & 0.25 & 0.19 & 0.32 & 0.25 \\
\hline
+ Graph Desc. & 0.28 & 0.35 & 0.34 & 0.48 & 0.24 \\
\hline
+ Physicochemical & 0.33 & 0.76 & 0.53 & 0.80 & 0.85 \\
\hline
Ensemble (GB) & 0.60 & \textbf{0.91} & 0.40 & \textbf{0.89} & \textbf{0.91} \\
\hline
Lasso Selection & 0.31 & 0.76 & \textbf{0.53} & 0.80 & 0.85 \\
\hline
Hybrid ($D/\zeta$ + Morgan) & \textbf{0.71} & 0.86 & 0.24 & 0.78 & 0.71 \\
\hline
\end{tabular}
\caption{$R^2$ values for all enhancement approaches across the five benchmark datasets. Bold indicates the best model per dataset. All values are mean $R^2$ from 5-fold cross-validation.}
\label{tab:results_all}
\end{table}

Several patterns emerge from these results. First, Ridge regularization produced no meaningful improvement on any dataset, confirming that the poor baseline performance comes from insufficient features rather than the regression algorithm. Second, adding physicochemical properties consistently produced the largest single improvement, particularly on SAMPL ($R^2$ from 0.24 to 0.85) and LogP synthetic ($R^2$ from 0.35 to 0.76). This indicates that chemical information matters more than graph topology alone for property prediction. Third, no single approach won on all datasets: Ensemble methods achieved the best results on three datasets (LogP synthetic, ESOL, SAMPL), the Hybrid approach was best for BACE, and Lasso was best for LogP experimental.

\begin{figure}[ht!]
    \centering
    \begin{subfigure}[b]{0.48\textwidth}
        \centering
        \includegraphics[width=\textwidth]{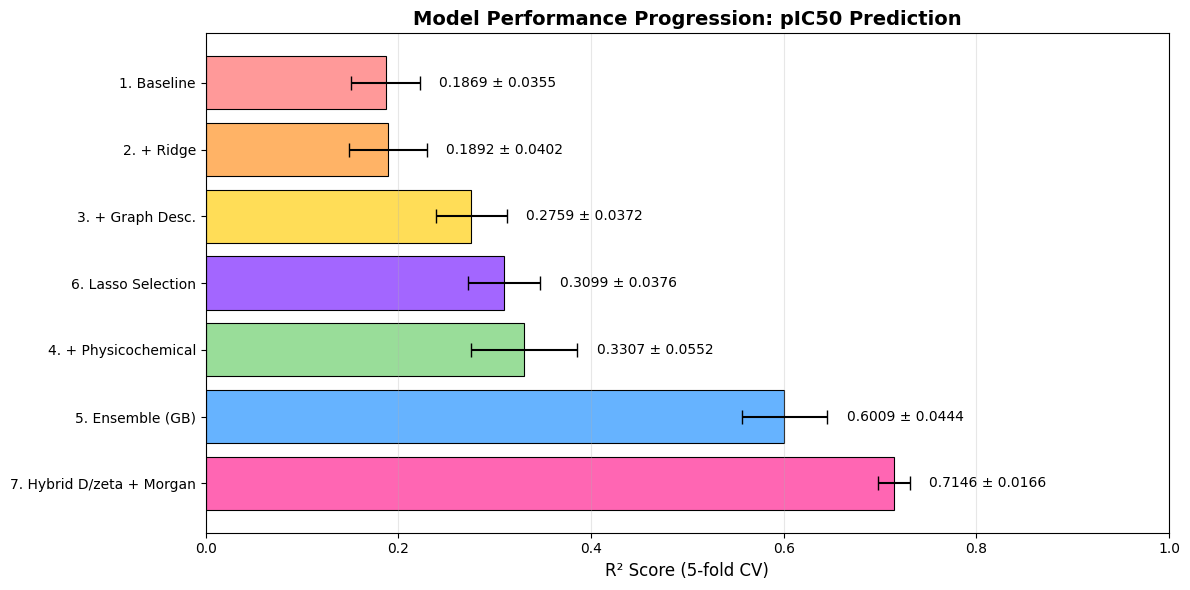}
        \caption{BACE dataset}
    \end{subfigure}
    \hfill
    \begin{subfigure}[b]{0.48\textwidth}
        \centering
        \includegraphics[width=\textwidth]{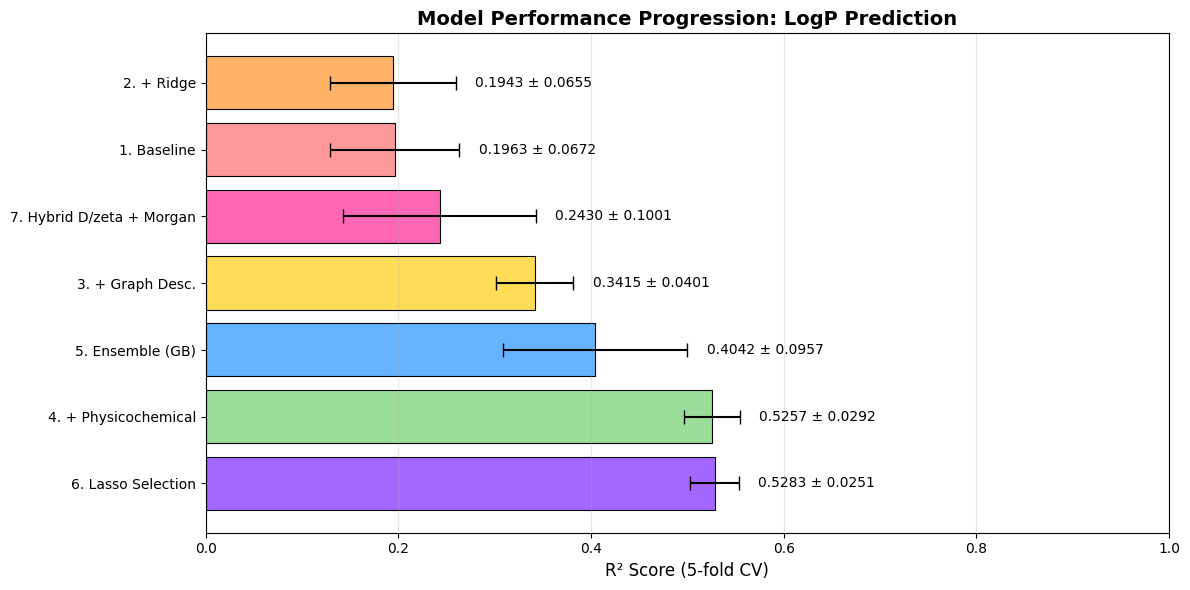}
        \caption{LogP Experimental dataset}
    \end{subfigure}

    \vspace{0.3cm}

    \begin{subfigure}[b]{0.48\textwidth}
        \centering
        \includegraphics[width=\textwidth]{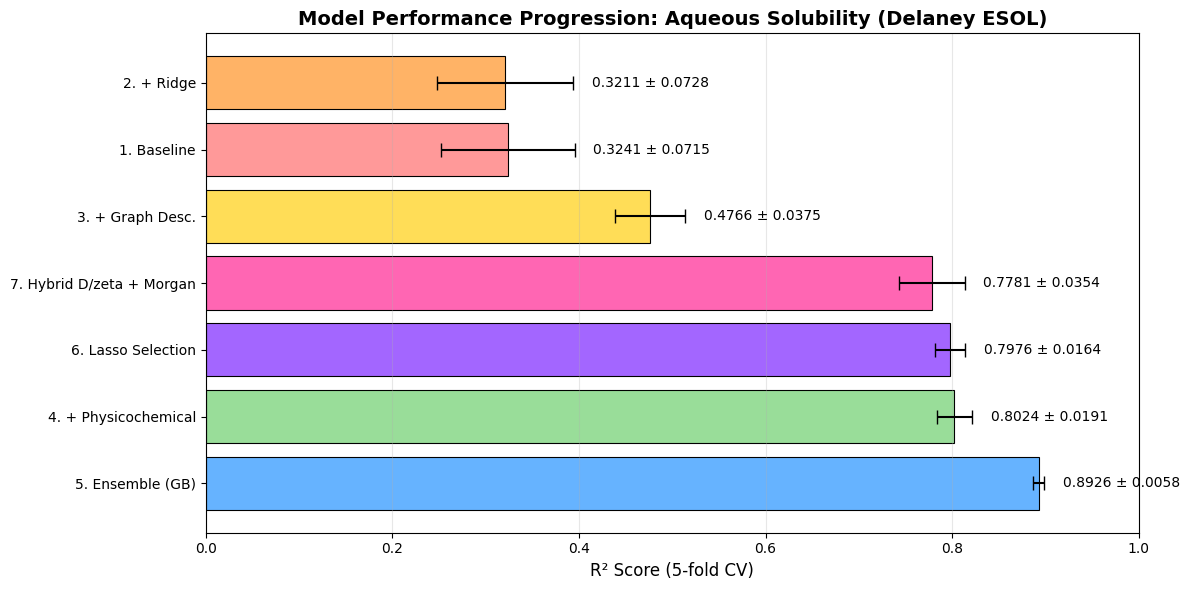}
        \caption{ESOL dataset}
    \end{subfigure}
    \hfill
    \begin{subfigure}[b]{0.48\textwidth}
        \centering
        \includegraphics[width=\textwidth]{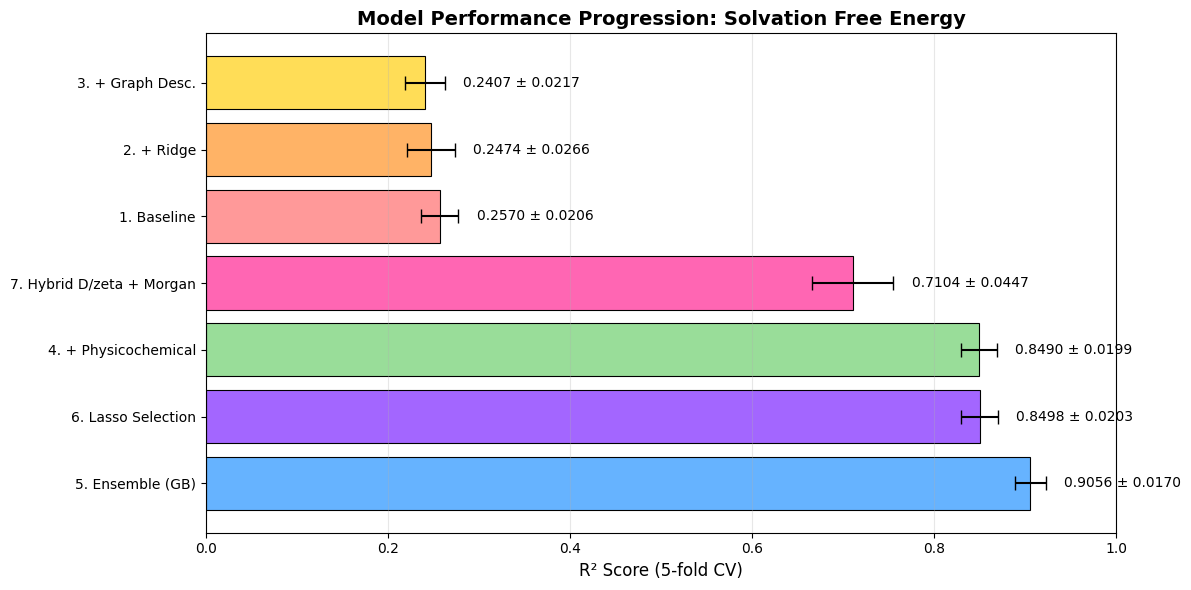}
        \caption{SAMPL dataset}
    \end{subfigure}
    \caption{$R^2$ performance progression across enhancement approaches for four benchmark datasets.}
    \label{fig:progression}
\end{figure}

Figure~\ref{fig:progression} shows the $R^2$ progression across enhancement approaches for four datasets, illustrating the consistent pattern where physicochemical properties and ensemble methods produce the largest performance gains.

Table~\ref{tab:best_summary} summarizes the best results and improvement percentages.

\begin{table}[ht!]
\centering
\begin{tabular}{|l|c|c|c|c|}
\hline
\textbf{Dataset} & \textbf{Baseline $R^2$} & \textbf{Best $R^2$} & \textbf{Best Model} & \textbf{Improvement} \\
\hline
BACE & 0.19 & 0.71 & Hybrid & 274\% \\
\hline
LogP Synthetic & 0.25 & 0.91 & Ensemble (GB) & 264\% \\
\hline
LogP Experimental & 0.20 & 0.53 & Lasso & 165\% \\
\hline
ESOL & 0.32 & 0.89 & Ensemble (GB) & 178\% \\
\hline
SAMPL & 0.26 & 0.91 & Ensemble (GB) & 250\% \\
\hline
\end{tabular}
\caption{Summary of best results per dataset with percentage improvement over the baseline.}
\label{tab:best_summary}
\end{table}

\subsubsection{Statistical significance}

Paired $t$-tests confirmed that all best models significantly outperform the baseline. Table~\ref{tab:stats} shows the results.

\begin{table}[ht!]
\centering
\begin{tabular}{|l|c|c|c|c|}
\hline
\textbf{Dataset} & \textbf{Baseline $R^2$} & \textbf{Best $R^2$} & \textbf{$t$-stat} & \textbf{$p$-value} \\
\hline
BACE & $0.19 \pm 0.04$ & $0.71 \pm 0.02$ & 31.11 & $<$0.0001 \\
\hline
LogP Synthetic & $0.25 \pm 0.01$ & $0.91 \pm 0.01$ & 150.78 & $<$0.0001 \\
\hline
LogP Experimental & $0.20 \pm 0.07$ & $0.53 \pm 0.03$ & 12.80 & 0.0002 \\
\hline
ESOL & $0.32 \pm 0.07$ & $0.89 \pm 0.01$ & 15.82 & $<$0.0001 \\
\hline
SAMPL & $0.26 \pm 0.02$ & $0.91 \pm 0.02$ & 67.06 & $<$0.0001 \\
\hline
\end{tabular}
\caption{Statistical significance of improvements. Values show mean $R^2 \pm$ 95\% confidence interval from 5-fold cross-validation.}
\label{tab:stats}
\end{table}

Ridge regularization showed no significant improvement on any dataset ($p > 0.05$ in all cases). Additional graph descriptors failed to produce a significant improvement on SAMPL ($p = 0.32$), and the hybrid approach was not significant on LogP experimental ($p = 0.09$), where the small dataset size (657 valid molecules) combined with 1,026 features leads to overfitting.

\subsubsection{Comparison with GCN}

\begin{figure}[h!]
    \centering
    \includegraphics[width=0.85\textwidth]{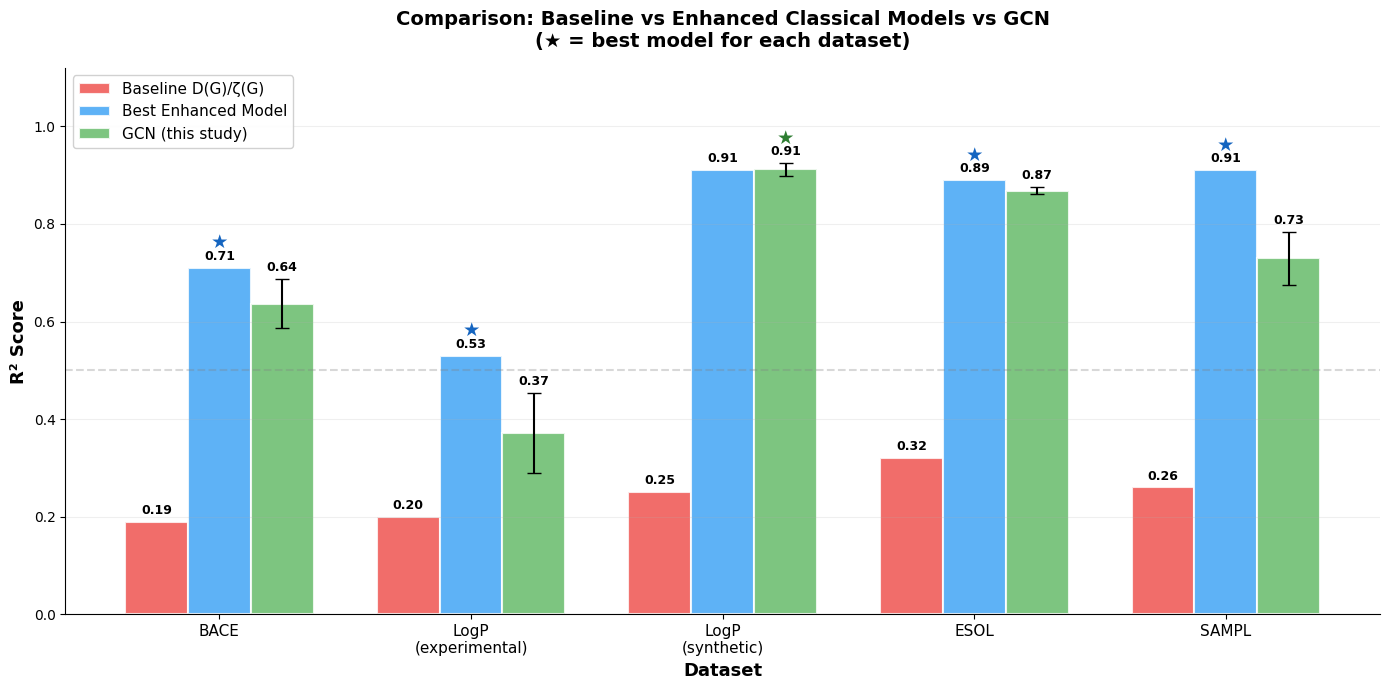}
    \caption{Comparison of baseline, best enhanced classical model, and GCN across all five benchmark datasets.} 
    \label{fig:gcn_comparison}
\end{figure}

Figure~\ref{fig:gcn_comparison} shows the direct comparison between the baseline, the best enhanced classical models, and the GCN across all five datasets.

Table~\ref{tab:gcn} presents the numerical comparison.

\begin{table}[ht!]
\centering
\begin{tabular}{|l|c|c|c|}
\hline
\textbf{Dataset} & \textbf{Best Enhanced $R^2$} & \textbf{GCN $R^2$} & \textbf{Winner} \\
\hline
BACE & $0.71 \pm 0.02$ & $0.64 \pm 0.05$ & Enhanced \\
\hline
LogP Synthetic & $0.91 \pm 0.01$ & $0.91 \pm 0.01$ & Tie \\
\hline
LogP Experimental & $0.53 \pm 0.03$ & $0.37 \pm 0.08$ & Enhanced \\
\hline
ESOL & $0.89 \pm 0.01$ & $0.87 \pm 0.01$ & Enhanced \\
\hline
SAMPL & $0.91 \pm 0.02$ & $0.73 \pm 0.05$ & Enhanced \\
\hline
\end{tabular}
\caption{Comparison between the best enhanced classical model and the GCN on each dataset. Both evaluated under identical conditions.}
\label{tab:gcn}
\end{table}

The enhanced classical models outperform or match the GCN on all five datasets. The largest gap is on SAMPL, where the ensemble model achieves $R^2 = 0.91$ compared to 0.73 for the GCN. On LogP synthetic, both approaches reach $R^2 = 0.91$.

\subsubsection{Comparison with Djagba et al.}

To place our results in a broader context, we compare with the recent study by Djagba et al.\ \cite{djagba2025hybrid}, who evaluated multiple deep learning architectures on the same datasets. Table~\ref{tab:djagba} presents this comparison.

\begin{table}[ht!]
\centering
\small
\setlength{\tabcolsep}{4pt}
\begin{tabular}{|l|c|c|c|c|c|}
\hline
\textbf{Dataset} & \textbf{This work} & \textbf{GNN+PGM} & \textbf{GNN} & \textbf{ChemBERTa} & \textbf{Winner} \\
\hline
BACE & 0.71 & \textbf{0.84} & 0.83 & 0.65 & DL \\
\hline
LogP Synth. & \textbf{0.91} & 0.82 & 0.82 & 0.95 & Ours \\
\hline
LogP Exp. & 0.53 & \textbf{0.66} & 0.63 & 0.39 & DL \\
\hline
ESOL & \textbf{0.89} & 0.86 & 0.84 & 0.82 & Ours \\
\hline
SAMPL & \textbf{0.91} & 0.91 & 0.91 & 0.89 & Tie \\
\hline
\end{tabular}
\caption{$R^2$ comparison with Djagba et al.\ \cite{djagba2025hybrid}. Note: this work uses 5-fold CV while Djagba et al.\ use an 80/20 split, so values are not directly comparable.}
\label{tab:djagba}
\end{table}

The enhanced classical models win on two datasets (ESOL and LogP synthetic), tie on one (SAMPL), and lose on two (BACE and LogP experimental). On ESOL, the ensemble model ($R^2 = 0.89$) outperforms the GNN+PGM ($R^2 = 0.86$) and all other deep learning approaches. On BACE, deep learning performs better ($R^2 = 0.84$ vs 0.71), which is expected since biological activity prediction involves complex structure-activity relationships that benefit from learned representations.

An important practical difference is that our entire framework trains in under five minutes on a standard laptop without GPU, while the GNN+PGM model of Djagba et al.\ requires an NVIDIA A100 GPU.

\subsubsection{Feature importance}

\begin{figure}[h!]
    \centering
    \includegraphics[width=0.75\textwidth]{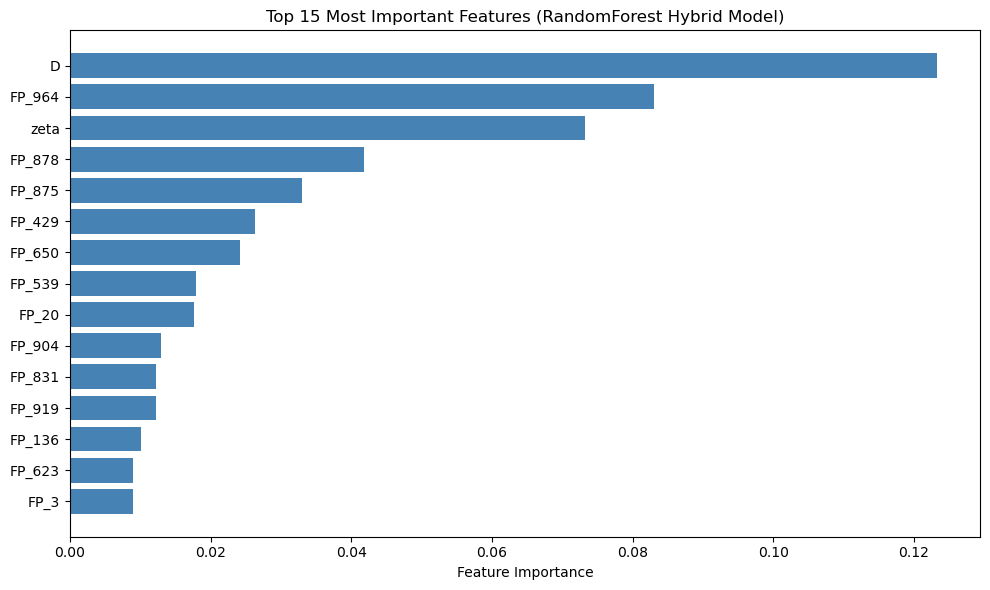}
    \caption{Top 15 most important features in the hybrid model for BACE prediction. }
    \label{fig:feature_importance}
\end{figure}

To understand which features drive the predictions, we examined feature importance in the best-performing hybrid model on the BACE dataset. Figure~\ref{fig:feature_importance} shows the top 15 most important features.

The external activity index $D(G)$ ranks as the single most important feature, followed by several Morgan fingerprint bits and the internal activity $\zeta(G)$. This result is significant because it confirms that the original graph-theoretic indices carry meaningful structural information even when combined with 1,024 fingerprint bits. The indices are not replaced by the fingerprints but rather complement them.

\subsection{Discussion}

The results reveal several important findings about the role of graph-theoretic models in molecular property prediction.

The baseline $D(G)$-$\zeta(G)$ model has limited generalization ability when applied beyond small homogeneous datasets. The degree-3 polynomial expansion of only two variables produces just 9 parameters and defines a smooth surface that cannot capture complex, non-smooth relationships between molecular structure and properties. Two global topological indices cannot distinguish molecules with different chemistry but similar graph structure, since atoms of different elements in the same position produce identical $D(G)$ and $\zeta(G)$ values. Furthermore, the $R^2 = 1$ of the original study was achieved with 9 molecules and 9 parameters, creating a 1:1 data-to-parameter ratio that carries a high risk of overfitting.

These limitations explain why each enhancement produces large improvements. Additional descriptors increase the input dimensionality, ensemble methods replace the rigid polynomial with flexible nonlinear models, and hybrid approaches add local structural details that are invisible to global indices.

A key finding is the strong dependence of optimal strategies on the target property and dataset characteristics. Biological activity prediction (BACE) benefits most from the hybrid fingerprint approach, which captures detailed local structural patterns important for protein-ligand interactions. Physicochemical properties (LogP, solubility, solvation energy) are better predicted by ensemble methods applied to physicochemical descriptors. For small datasets like LogP experimental (657 valid molecules), Lasso feature selection outperforms more complex approaches by removing irrelevant features that would otherwise cause overfitting.

The comparison with deep learning approaches supports the practical value of the enhancement framework. For physicochemical property prediction, enhanced classical models can match or exceed deep learning at a fraction of the computational cost. For biological activity prediction, deep learning retains an advantage due to its ability to learn complex representations from data. These findings can guide researchers in choosing the right approach based on their specific task and available resources.

\section{Conclusion}

This study evaluated and enhanced the Mukwembi-Nyabadza graph-theoretic model for molecular property prediction on five benchmark datasets from MoleculeNet. The baseline model, which uses only the external activity $D(G)$ and internal activity $\zeta(G)$ indices, achieved an average $R^2$ of 0.24, confirming limited transferability from small homogeneous datasets to large diverse molecular collections.

Through a systematic enhancement framework incorporating regularization, additional descriptors, physicochemical properties, ensemble learning, feature selection, and hybrid representations, the average best $R^2$ improved to 0.79. All improvements were statistically significant ($p < 0.001$ on all five datasets). The enhanced models matched or outperformed a GCN on all five datasets and achieved competitive results against the published deep learning models of Djagba et al., winning on two datasets and tying on one, despite requiring no GPU and training in under five minutes.

Several limitations should be noted. The $D(G)$ and $\zeta(G)$ indices capture only global topology, not three-dimensional geometry or electronic effects. Performance on small datasets remains moderate (LogP experimental $R^2 = 0.53$). The GCN comparison used a standard architecture, and more advanced GNN models could potentially achieve higher performance.

Future work could extend this framework to additional molecular properties such as pKa and membrane permeability. Applying the methodology to molecular targets relevant to African public health priorities, such as antimalarial or anti-tuberculosis compounds, would demonstrate its value for drug discovery in resource-limited settings. The entire codebase is publicly available on GitHub, enabling other researchers to reproduce and extend this work.

\newpage

\bibliography{biblio}
\bibliographystyle{plain}
\end{document}